\documentclass{article}
\usepackage{spconf, amsmath, graphicx, multirow, array, color}

\title{Richly Activated Graph Convolutional Network for Action Recognition with Incomplete Skeletons}

\name{Yi-Fan Song, Zhang Zhang and Liang Wang\thanks{This work is sponsored by National Key R\&D Program of China (2016YFB1001002), National Natural Science Foundation of China (61525306, 61633021, 61721004, 61420106015) and CAS-AIR.}}
\address{
School of Artificial Intelligence, University of Chinese Academy of Sciences (UCAS), Beijing, China \\
Institute of Automation, Chinese Academy of Sciences (CASIA), Beijing, China \\
Center for Research on Intelligent Perception and Computing (CRIPAC), Beijing, China \\
}

\begin{document}

\maketitle

\begin{abstract}
Current methods for skeleton-based human action recognition usually work with completely observed skeletons. However, in real scenarios, it is prone to capture incomplete and noisy skeletons, which will deteriorate the performance of traditional models. To enhance the robustness of action recognition models to incomplete skeletons, we propose a multi-stream graph convolutional network (GCN) for exploring sufficient discriminative features distributed over all skeleton joints. Here, each stream of the network is only responsible for learning features from currently unactivated joints, which are distinguished by the class activation maps (CAM) obtained by preceding streams, so that the activated joints of the proposed method are obviously more than traditional methods. Thus, the proposed method is termed richly activated GCN (RA-GCN), where the richly discovered features will improve the robustness of the model. Compared to the state-of-the-art methods, the RA-GCN achieves comparable performance on the NTU RGB+D dataset. Moreover, on a synthetic occlusion dataset, the performance deterioration can be alleviated by the RA-GCN significantly.
\end{abstract}


\section{Introduction}
\label{sec:intro}

\begin{figure}[t]
\label{fig:1}
\centerline{\includegraphics[width=9.8cm]{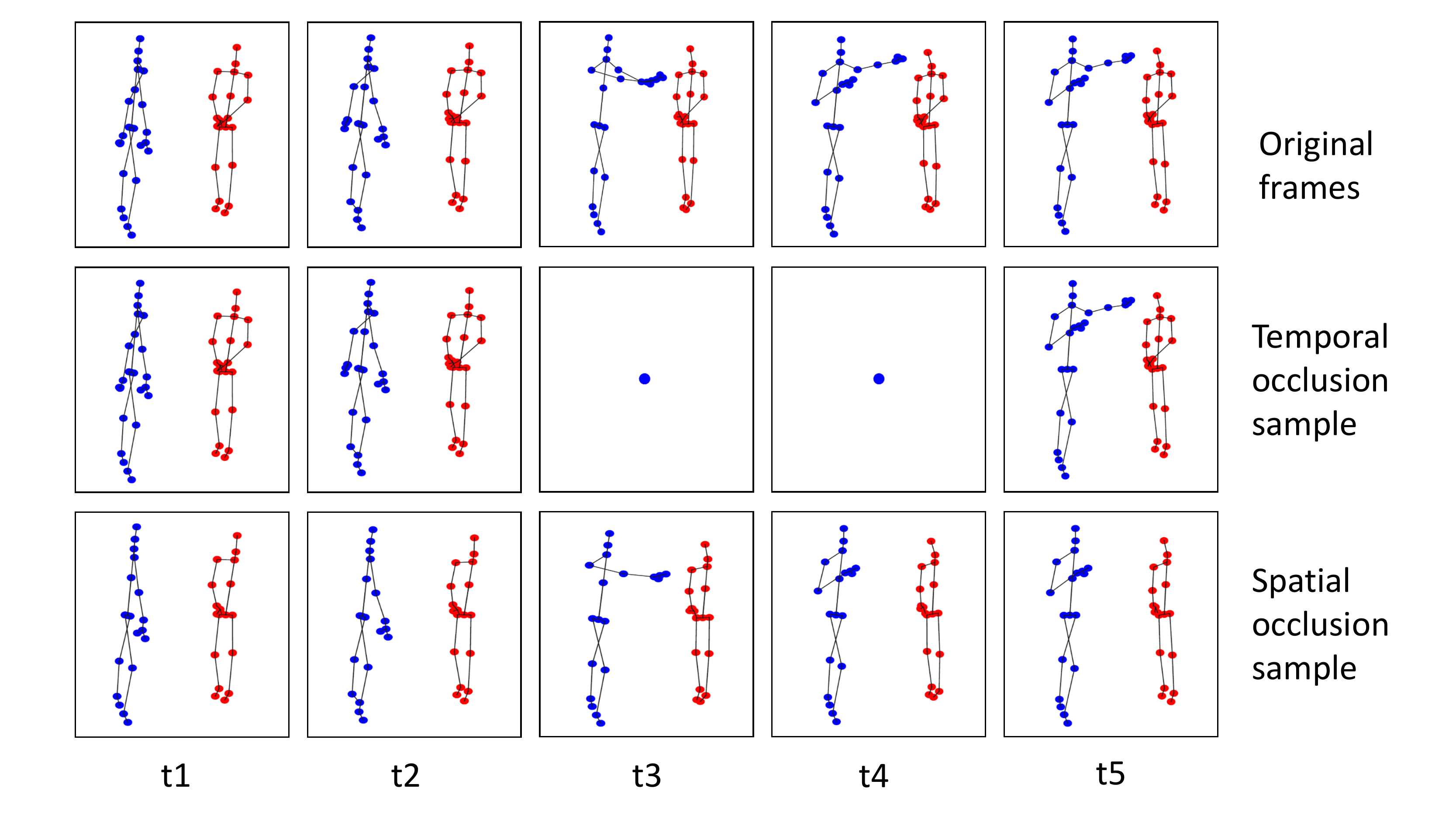}}
\caption{The demonstration of the occlusion dataset based on the NTU RGB+D dataset.}
\vspace{-0.2cm}
\end{figure}

Skeleton-based human action recognition methods become increasingly important in many applications and achieve great progress, due to the superiority in background adaptability, robustness to light intensity and less computational cost. Skeleton data is composed of 3D coordinates of multiple spatial and temporal skeleton joints, which can be either collected by multimodal sensors such as Kinect or directly estimated from 2D images by pose estimation methods. Traditional methods usually deal with skeleton data in two ways. One way is to connect these joints into a whole vector, then model temporal information using RNN-based methods \cite{Liu2016, Zhang2017, Zhang2017b, Wang2017, Si2018}. The other way is to treat or expand temporal sequences of joints into images, then utilize CNN-based methods to recognize actions \cite{Li2017b, Kim2017, Ding2017, Li2018, Tang2018}. However, the spatial structure information among skeleton joints is hard to be ultilized effectively by both the RNN-based and CNN-based methods, though researchers propose some additional constraints or dedicated network structures to strenuously encode the spatial structure of skeleton joints. Recently, Yan et al.\cite{Yan2018} firstly apply graph-based methods to skeleton-based action recognition, and propose the spatial temporal graph convolutional networks (ST-GCN) to extract features embedded in the spatial configuration and the temporal dynamics. Additionally, there are also some action recognition methods using graphic techniques, such as \cite{Wang2013, Wang2014a, Wang2014b}. But these are mainly based on RGB videos, instead of skeleton data.

\begin{figure*}[t]
\label{fig:2}
\centerline{\includegraphics[width=16cm]{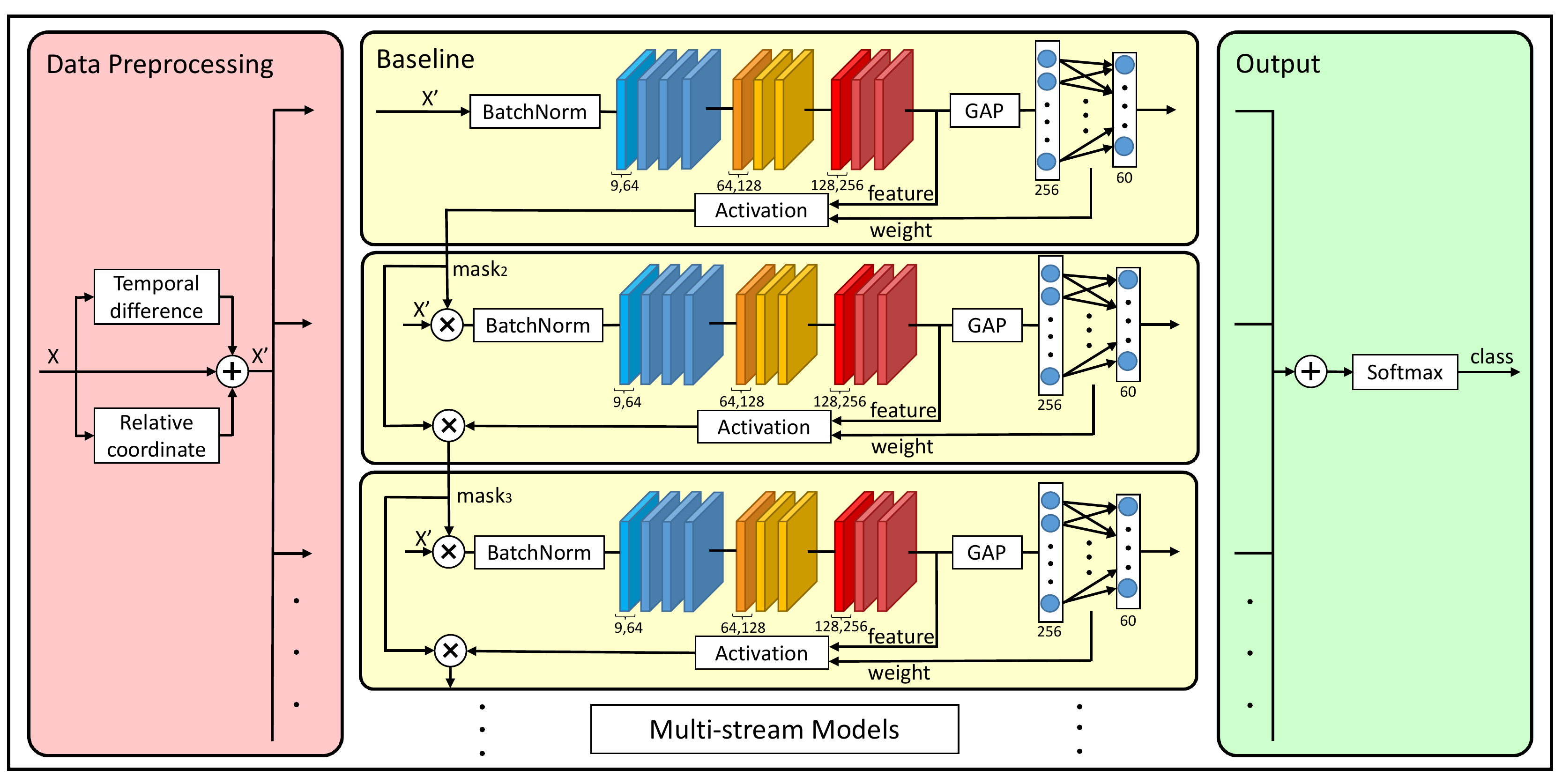}}
\caption{The pipeline of RA-GCN with three baseline models. Each baseline model is an ST-GCN network with ten layers. The two numbers under the ST-GCN layers are input channels and output channels, respectively. Other layers contain the same input and output channels. Each layer with different input and output channels uses a temporal stride 2 to reduce the sequence length. GAP is global average pooling operation, and $\otimes$ and $\oplus$ denote element-wise multiplication and concatenation, respectively.}
\vspace{-0.2cm}
\end{figure*}

All these above methods assume that the complete skeleton joints can be well captured, while the incomplete case is not considered. However, it is often difficult to obtain a complete skeleton sequence in real scenarios. For example, pedestrians may be occluded by parked vehicles or other contextual objects observed. Meanwhile, when facing to incomplete skeletons, traditional methods will have varying degrees of performance deterioration. Therefore, how to recognize actions with incomplete skeletons is a challenging problem.

Many researchers are exploring to extract non-local features from all positions in the input feature maps, such as \cite{Wang2018} and \cite{Li2018c}. Inspired by this, we propose a multi-stream graph convolutional network (GCN) to explore sufficient discriminative features for robust action recognitions. Here, a subtle technique, class activation maps (CAM), is utilized to distinguish the discriminative skeleton joints activated by each stream. The activation maps obtained by preceding streams are accumulated as a mask matrix to inform the new stream about which joints have been already activated. Then, the new stream will be forced to explore discriminative features from unactivated joints. Therefore, the proposed method is called richly activated GCN (RA-GCN), where the richly discovered discriminative features will improve the robustness of the model to incomplete skeletons. The experimental results on the NTU RGB+D dataset \cite{Shahroudy2016} show that the RA-GCN achieves comparable performance to the state-of-the-art methods. Furthermore, for the case of incomplete skeletons, we construct a synthetic occlusion dataset, where the joints in the NTU dataset are partially occluded over both spatial and temporal dimensions. Some examples on the two types of occlusions are shown in Fig.1. On the new dataset, the RA-GCN significantly alleviates the performance deterioration.

\section{Richly Activated GCN}
\label{sec:RAGCN}

In order to enhance the robustness of action recognition models, we propose the RA-GCN to explore sufficient discriminative features from the training skeleton sequences. The overview of RA-GCN is presented in Fig.2. Suppose that $V$ is the number of joints in one skeleton, $M$ is the number of skeletons in one frame and $T$ is the number of frames in one sequence. Then, the size of input data $\bf x$ is $C\times T\times V\times M$, where $C=3$ denotes the 3D coordinates of each joint. 

The proposed network consists of three main steps. Firstly, in the preprocessing module, for extracting more informative features, the input data $\bf x$ is transformed into $\bf x'$, whose size is $3C \times T\times V\times M$. The preprocessing module is composed of two parts. The first part extracts motion features by computing temporal difference ${\bf x}_t={\bf x}[t+1]-{\bf x}[t]$, where ${\bf x}[t]$ means the input data of the $t$-th frame. The second part calculates the relative coordinates ${\bf x}_r$ between all joints and the center joint (center trunk) in each frame. Then, $\bf x'$ will be obtained by concatenating $\bf x$, ${\bf x}_t$ and ${\bf x}_r$. Secondly, for each stream, the skeleton joints in $\bf x'$ will be filtered by the element-wise product with a mask matrix, which records the currently unactivated joints. These joints are distinguished by accumulating the activated maps obtained by the activation modules of preceding streams. Here, the mask matrix is initialized to all-one matrix with the same size as $\bf x'$. After the masking operation, the input data of each stream only contains unactivated joints, and subsequently passes through an ST-GCN network \cite{Yan2018} to obtain a feature representation based on partial skeleton joints. Finally, the features of all streams are concatenated in the output module, and a softmax layer is used to obtain the final class of input $\bf x$.

\subsection{Baseline Model}
\label{ssec:baseline}

The baseline model is the ST-GCN \cite{Yan2018}, which is composed of several spatial convolutional blocks and temporal convolutional blocks. Concretely, the spatial graph convolutional block can be implemented by the following formulation:
\begin{equation}
{\bf f}_{out} = \sum_{d=0}^{D_{max}} {\bf W}_d{\bf f}_{in} ({\bf \Lambda}_d^{-\frac{1}{2}}{\bf A}_d{\bf \Lambda}_d^{-\frac{1}{2}}\otimes {\bf M}_d),
\end{equation}
where $D_{max}$ is the predefined maximum distance, ${\bf f}_{in}$ and ${\bf f}_{out}$ are the input and output feature maps respectively, ${\bf A}_d$ denotes the adjacency matrix for graphic distance $d$, $\Lambda_d^{ii} = \sum_k A_d^{ik} + \alpha$ is the normalized diagonal matrix, $A_d^{ik}$ denotes the element of the $i$-th row and $k$-th column of ${\bf A}_d$ and $\alpha$ is set to a small value, e.g. $10^{-4}$, to avoid the empty rows in ${\bf \Lambda}_d$. For each adjacency matrix, we accompany it with a learnable matrix ${\bf M}_d$, which expresses the importance of each edge.

After the spatial graph convolutional block, a $1\times L$ Conv layer is used to extract temporal information of the feature map ${\bf f}_{out}$, where $L$ is the temporal window size. Both spatial and temporal convolutional blocks are followed with a BatchNorm layer and a ReLU layer, and the total ST-GCN layer contains a residual connection. Besides, an adaptive dropout layer is added between the spatial and temporal convolutional blocks to avoid overfitting. More details of the ST-GCN will be found in \cite{Yan2018}.

\subsection{Activation Module}
\label{ssec:activation}

The activation module in the RA-GCN is constructed to distinguish the activated joints of each stream, then guide the learning process of the new stream by accumulating the activated maps of preceding streams. This procedure can be implemented mainly by the CAM technique \cite{Zhou2016}. The original CAM technique is to localize class-specific image regions in a single forward-pass, and $M_c$ is defined as the activation map for class $c$, where each spatial point is
\begin{equation}
M_c(x,y)=\sum_k w_k^c f_k(x,y).
\end{equation}
In this formulation, $f_k(x,y)$ is the feature map before the global average pooling operation, and $w_k^c$ is the weight of the $k$-th channel for class $c$. In this paper, we replace the coordinates $(x,y)$ in an image with the frame number $t$ and the joint number $i$ in a skeleton sequence, by which we are able to locate the activated joints. These joints can also be regarded as the attention joints of the corresponding streams. Here, the class $c$ is selected as the true class. Then, the mask matrix of stream $s$ is calculated as follows.
\begin{equation}
mask_s=(\prod_{i=1}^{s-1}mask_i)\otimes(1-Softmax(M_c^{s-1})),
\end{equation}
where $\prod$ denotes the element-wise product of all mask matrices before the $s$-th stream. Specially, the mask matrix of the first stream is an all-one matrix. Finally, the input of stream $s$ will be obtained by
\begin{equation}
\label{Eq:4}
{\bf x}_s={\bf x'}\otimes mask_s,
\end{equation}
where $\bf x'$ is the skeleton data after preprocessing. 

Eq.~\ref{Eq:4} illustrates that the input of stream $s$ only consists of the joints which are not activated by previous streams. Thus the RA-GCN will explore discriminative features from all joints sufficiently. Fig.3 shows an example of the activated joints of all streams for the baseline model, the RA-GCN whth 2 streams and 3 streams, from which we can observe that the RA-GCN discovers significantly more activated joints than the baseline model. The code of RA-GCN is availabel on \texttt{https://github.com/yfsong0709/RA-GCNv1}

\begin{figure}[t]
\label{fig:3}
\centerline{\includegraphics[width=9.8cm]{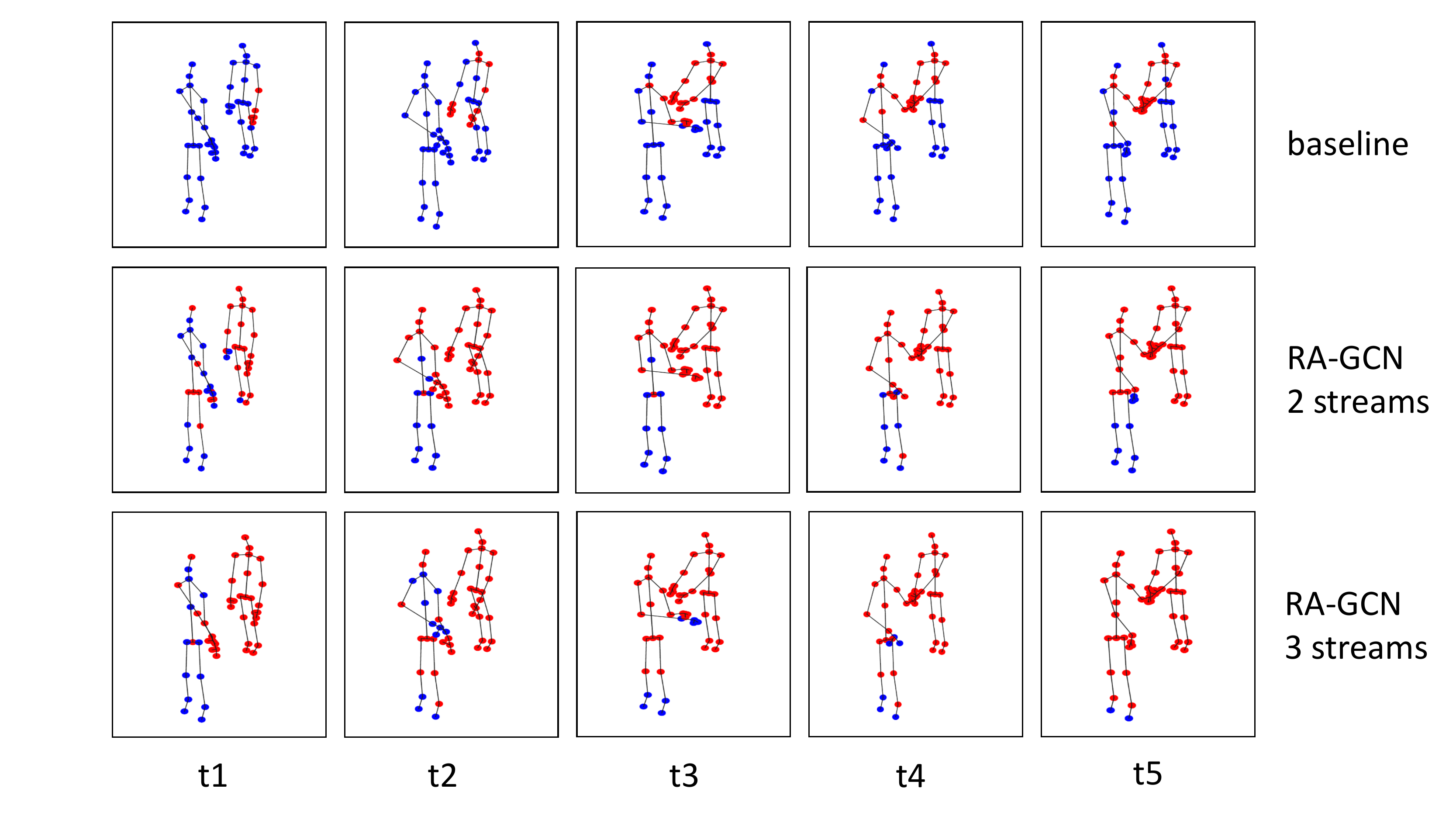}}
\caption{An example of activated joints of all streams for the baseline model, the RA-GCN whth 2 streams and 3 streams. The {\color{red}red} points denote the activated joints, while the {\color{blue}blue} points denote the unactivated joints. Best viewed in color.}
\vspace{-0.2cm}
\end{figure}

\section{Experiments}
\label{sec:experiments}

\subsection{Dataset and Implemental Details}
\label{ssec:dataset}

In this section, we evaluate the performance of the RA-GCN on a large-scale dataset {\bf NTU RGB+D} \cite{Shahroudy2016}, which is the currently largest indoor action recognition dataset. This dataset contains 56880 video samples collected by Microsoft Kinect v2, and consists of 60 action classes performed by 40 subjects. The maximum frame number $T$ is set to 300 for simplicity. The authors of this dataset recommend two benchmarks:  (1) {\bf cross-subject} (CS) contains 40320 and 16560 samples for training and evaluation, which splits 40 subjects into training and evaluation groups; (2) {\bf cross-view} (CV) contains 37920 and 18960 samples, which uses camera 2 and 3 for training and camera 1 for evaluation.

Before training RA-GCN, we need to pre-train an ST-GCN network with transformed skeleton data to get the baseline model, and the training setting is the same as \cite{Yan2018}. Then, initialize all streams of RA-GCN with this baseline model. Finally, finetune the RA-GCN model. All of our experiments are running on two TITAN X GPUs.

\begin{table}
\label{tab:1}
\caption{Comparison of methods on NTU RGB+D (\%)}
\centering
\begin{tabular}{cccc}
\hline
& year & CS & CV \\
\hline
H-BRNN \cite{Du2015} & 2015 & 59.1 & 64.0 \\
PA-LSTM \cite{Shahroudy2016} & 2016 & 62.9 & 70.3 \\
VA-LSTM \cite{Zhang2017b} & 2017 & 79.4 & 87.6 \\
ST-GCN (baseline) \cite{Yan2018} & 2018 & 81.5 & 88.3 \\
HCN \cite{Li2018} & 2018 & 86.5 & 91.1 \\
SR-TSL \cite{Si2018} & 2018 & 84.8 & 92.4 \\
PB-GCN \cite{Thakkar2018} & 2018 & {\bf 87.5} & 93.2 \\
\hline
*2s RA-GCN & -- & 85.8 & 93.0 \\
*3s RA-GCN & -- & 85.9 & {\bf 93.5} \\
\hline
\multicolumn{4}{l}{*: 2s denotes two streams and 3s denotes three streams}
\end{tabular}
\end{table}

\begin{table}
\label{tab:2}
\caption{Comparison of different maximum distances and temporal window sizes on NTU RGB+D (\%)}
\centering
\begin{tabular}{ccccc}
\hline
& & CS &  & CV \\
\hline
$D_{max}=1, L=5$&  & 85.2 & &  90.5 \\
$D_{max}=2, L=5$&  & {\bf 85.9} & & 91.6 \\
$D_{max}=3, L=5$&  & 85.8 & & 92.2 \\
$D_{max}=1, L=9$&  & 85.2 & & 91.7 \\
$D_{max}=2, L=9$&  & 85.4 & & 92.7 \\
$D_{max}=3, L=9$&  & 85.0 & & {\bf 93.5} \\
\hline
\end{tabular}
\end{table}

\subsection{Experimental results on complete skeletons}
\label{ssec:results}

We compare the performance of RA-GCN against previous state-of-the-art methods on the NTU RGB+D dataset. As shown in Table 1, our method achieves better performance than other models on the CV benchmark, though it is only 1.6\% less than PB-GCN \cite{Thakkar2018} on the CS benchmark. Compared to the baseline, our method outperforms by 4.4\% and 5.2\%, respectively. The RA-GCN only achieves comparable performance to the state-of-the-art method, because the RA-GCN aims to discover more discriminative joints, while the most actions can be recognized by only a few main joints. However, when these main joints are occluded, the performance of traditional methods will deteriorate significantly.

\subsection{Ablation studies}
\label{ssec:ablation}

In Section~\ref{ssec:baseline}, we introduce two hyperparameters for the baseline model, $D_{max}$ for maximum distance and $L$ for temporal window size. These two hyperparameters have a great impact on our model. We test six groups of parameters and the experimental results are given in Table 2. It is observed that our model achieves the best accuracy when $D_{max}=2$ and $L=5$ on the CS benchmark. As to the CV benchmark, $D_{max}$ and $L$ are optimally set to 3 and 9, respectively.

\begin{table}
\label{tab:3}
\caption{Occlusion experiments on the CS benchmark (\%)}
\centering
\begin{tabular}{ccccccc}
\hline
spatial & \multicolumn{6}{c}{occluded part} \\
\cline{2-7}
occlusion & none & 1 & 2 & 3 & 4 & 5 \\
\hline
baseline \cite{Yan2018} & 80.7 & 71.4 & {\bf 60.5} & 62.6 & 77.4 & 50.2 \\
SR-TSL \cite{Si2018} & 84.8 & 70.6 & 54.3 & 48.6 & 74.3 & 56.2 \\
2s RA-GCN & 85.8 & 72.8 & 58.3 & 73.2 & 80.3 & 70.6 \\
3s RA-GCN & {\bf 85.9} & {\bf 73.4} & 60.4 & {\bf 73.5} & {\bf 81.1} & {\bf 70.6} \\
*difference & 5.2 & 2.0 & -0.1 & 10.9 & 3.7 & 20.4 \\
\hline
\hline
temporal & \multicolumn{6}{c}{occluded frame number} \\
\cline{2-7}
occlusion & 0 & 10 & 20 & 30 & 40 & 50 \\
\hline
baseline \cite{Yan2018} & 80.7 & 69.3 & 57.0 & 44.5 & 34.5 & 24.0 \\
SR-TSL \cite{Si2018} & 84.8 & 70.9 & 62.6 & 48.8 & 41.3 & 28.8 \\
2s RA-GCN & 85.8 & {\bf 82.0} & 74.7 & 64.9 & 52.5 & 38.6 \\
3s RA-GCN & {\bf 85.9} & 81.9 & {\bf 75.0} & {\bf 66.3} & {\bf 54.4} & {\bf 40.6} \\
*difference & 5.2 & 12.6 & 18.0 & 21.8 & 19.9 & 16.6 \\
\hline
\multicolumn{7}{l}{*: the difference between 3s RA-GCN and baseline model}
\end{tabular}
\end{table}

\subsection{Experimental results on incomplete skeletons}
\label{ssec:occlusion}

To validate the robustness of our method to incomplete skeletons, we construct a synthetic occlusion dataset based on the NTU-RGB+D dataset, where some joints are selected to be occluded (set to 0) over both spatial and temporal dimensions. For spatial occlusion, we train the testing models with complete skeletons, then evaluate them with skeletons without part 1, 2, 3, 4, 5, which denote left arm, right arm, two hands, two legs and trunk, respectively. For temporal occlusion, we randomly occlude a block of frames in first 100 frames, because the lengths of many sequences are less than 100. Some examples of the occlusion dataset are demonstrated in Fig.1. On the synthetic occlusion dataset, we test the baseline model \cite{Yan2018}, SR-TSL \cite{Si2018}, RA-GCN with 2 streams and 3 streams. The experimental results are displayed in Table 3, from which it is revealed that the 3s RA-GCN greatly outperforms the other models in most occlusion experiments except occluding part 2. Especially, in temporal occlusion experiments, 3s RA-GCN achieves an increasing superiority to other models. Thus the performance deterioration can be alleviated by the proposed methods. We also find that when some important joints, such as right arms, are occluded, some action categories, e.g. handshaking, cannot be inferred by other joints. The proposed method will fail in such case.

\section{Conclusion}
\label{sec:conclusion}

In this paper, we have proposed a novel model named RA-GCN, which achieves much better than the baseline model and improves the robustness of the model. With extensive experiments on the NTU RGB+D dataset, we verify the effectiveness of our model in occlusion scenarios.

In the future, we will add the attention module into our model, in order to make each stream focus more on certain discriminative joints.

\bibliographystyle{IEEEbib}
\bibliography{refs}

\end{document}